\documentclass[sigconf]{acmart}
\usepackage{booktabs} % For formal tables

\usepackage[utf8]{inputenc}
\usepackage{graphicx}

\usepackage{xcolor}
\usepackage{mdframed}

\setcopyright{rightsretained}
\acmDOI{}

\acmISBN{}

\acmConference{ICTIR' 17 Workshop on Search-Oriented Conversational AI (SCAI' 2017)}{October 1, 2017}{Amsterdam, Netherlands}
\acmYear{}
\copyrightyear{2017}

\acmPrice{}

\begin{document}
\title[WOAH Methodology]{WOAH: Preliminaries to Zero-shot Ontology Learning for Conversational Agents}
%\titlenote{Produces the permission block, and copyright information}
%\subtitle{Extended Abstract}
%\subtitlenote{The full version of the author's guide is available as \texttt{acmart.pdf} document}

\author{Gonzalo Estrán Buyo}
%\authornote{}
%\orcid{}
\affiliation{%
  \institution{Telefónica R\&D and Universidad Politécnica de Madrid}
  %\streetaddress{}
  \city{Madrid} 
  \state{Spain} 
  %\postcode{}
}
\email{gonzalo.esbuyo@alumnos.upm.es}

\begin{abstract}
The present paper presents the Weighted Ontology Approximation Heuristic (WOAH), a novel zero-shot approach to ontology estimation for conversational agents development environments.
This methodology extracts verbs and nouns separately from data by distilling the dependencies obtained and applying similarity and sparsity metrics to generate an ontology estimation configurable in terms of the level of generalization.
%\footnote{This is an abstract footnote}
\end{abstract}

\maketitle

\section{Introduction}
Current methods of design for conversational agents consist mainly in the definition of the entities (object types) and intents (possible actions with those entities) that are pertinent for the services that it wants to offer. To structure and define those entities and intents, an appropriate ontology design is needed in order to specify which actions are possible for which objects. Sometimes those relations can be very difficult to determine, not only because of the number of possibilities to consider, but also due to the fact that not all the intents and entities are equally related (it is more likely to perform some actions than others with an object).

If the word \textit{bank} is considered, it may refer to an organization, a sloping raised land, or a mass of something depending on the context referred. If your bank wants to design a conversational agent, it probably will not want to consider more meanings than the first one, and in any case it will want to assign more importance to that meaning if it considers others. The problem comes from the \textit{a priori} knowledge applied by the entity recognizer tools in order to classify words, which are very useful for general purpose conversational agents but are too generic for specific purpose ones. This observed ambiguities can be seen as uncertainties that are introduced by the recognizer and were not present in the considered domain. From this conflict, it can be asserted that entity detection techniques consider some contexts that are not relevant for a restricted domain.

Furthermore, applied methods of word representation through vectors do not appropriately consider the semantics, such as bag-of-words \cite{harris}, or obtain dimensions which are not descriptive of the domain where those words belong, such as \textit{word2vec} \cite{mikolov14}. Those representation models have another defect when they are considered for conversational agents: they do not differentiate between verbs and nouns.

This paper presents the WOAH (Weighted Ontology Approximation Heuristic) methodology, which is intended to extract the words from a dataset differentiating between verbs (actions that then are abstracted into intents) and nouns (objects then abstracted into entities). It then represents those verbs and words separately as embeddings with dimensions that are descriptive of the data domain in order to obtain an ontological approximation without previous knowledge about the data. This approximation also permits different levels of abstraction and provides the level of associativity between extracted entity and intent estimations. Finally, the level of generality for the predicted ontology can be adjusted to provide a general to more specific range of possible estimations.

\section{Related Work}

A wide range of studies have been done on Ontology Learning, a field from the mature area of Ontology Engineering. The ontology that wants to be obtained can be classified in different ways based on the following aspects \cite{upm}:
\begin{enumerate}
\item Based on the \textit{richness of the internal structure}: in this case an \textbf{informal “is-a” hierarchy}. The reason is that the structure to capture is focused on relating entities under the same intent instead of specifying which entities are instances of others.
\item Based on the \textit{subject of the conceptualization}: in our case the ontology to obtain will be a \textbf{Domain-Task ontology}. More precisely, an ontology that describes the vocabulary related to a generic task or activity and will be reusable in a given domain, but not across domains.
\end{enumerate} 

Usual components of an ontology learning (OL) process \cite{handbook} consist in an \textbf{ontology management} to provide an interface between the ontology and the learning algorithms (to successfully add new concepts, relations or axioms), a \textbf{coordination} to sequentially arrange and apply the algorithms selected by the user (passing
the results to each other), a \textbf{resource processing} to discover, import, analyze and transform data and an \textbf{algorithm library component} with customized versions of the necessary machine learning algorithms and a comprehensive
number of implemented distance or similarity measures.

Furthermore, common approaches to OL include:
\begin{enumerate}
\item \textbf{hierarchical clustering}, where sets of terms are organized in a hierarchy that can be transformed directly into a prototype-based ontology.
\item \textbf{distributional similarity methods}: these can be \textbf{syntactic} if make use of similarity regarding predicate-argument relations
(i.e. verb-subject and verb-object relations), or \textbf{window-based} if rate pairs of word high that occur
together often in a certain text window without previous preprocessing.
\item \textbf{extraction pattern methods}, using explicit clues with lexico-syntactic patterns.
\item Treating OL as a \textbf{classification task} where features of the existing data are used as a training set for Machine Learning, which produces a
classifier for previously unknown instances.
\item Learning the ontology through \textbf{semantic lexicon construction}, but generating a more fine-grained encode of the semantic similarities
between terms and then abstracting terms to concepts.
\end{enumerate}
\textit{OntoLearn Reloaded} \cite{ontolearn} is a well-known example of an accurate (graph-based) methodology to learn ontologies. It works extracting an initial terminology and its hypernyms, which then are used to filter the domain and produce an optimal tree-like taxonomy of the initial noisy graph.

However, these OL techniques do not focus on domain-task ontologies for conversational agents so, instead of trying to extract intents (from verbs) and associate the learned entities (from objects) to those, they use the verbs in order to extract more accurately the entities and their relations. In other words, those approaches are oriented to entities more than to intents.

On the other hand, some proposals have been presented in the last years to make use of ontologies or types to extract better the semantic relationships and representations (such as \cite{ontologyaware}, \cite{types} or \cite{nasari}), but they were not focused in the division between entities and intents and also were considerably dependent of \textit{a priori} data.

\section{Methodology}

WOAH is based on the idea that \textbf{words do not have an intrinsic meaning but only contexts}. A \textbf{meaning} can be understood as \textit{one of the possible contextualizations of a word}. And a \textbf{contextualization} can be defined as the \textit{use of a word} through two possible operations:
\begin{itemize}
\item \textbf{association} of the word with another word.
\item \textbf{action} with or over that word.
\end{itemize}
In other terms, when we use words is when they have a meaning, and we use them through sentences in which we relate those words with others to transmit a specific idea.

The inherent idea is to extract the words that should be considered relevant, and for those words obtain a stochastic measure of the possible associations and actions. 

The result consists of matrix of actions ($I$) with intents as elements and objects as dimensions, and a matrix of associations ($E$) with entities as elements and complements as dimensions. The adjacencies of those two matrices can the be converted into an ontology with the needed services (entities) and actions (intents).

To achieve that, the WOAH is divided into successive phases. As \textit{Figure 1} shows, these phases range from the preprocessing to the generation of the resulting approximated ontology.

\begin{figure}
  \centering
    \includegraphics[width=0.5\textwidth]{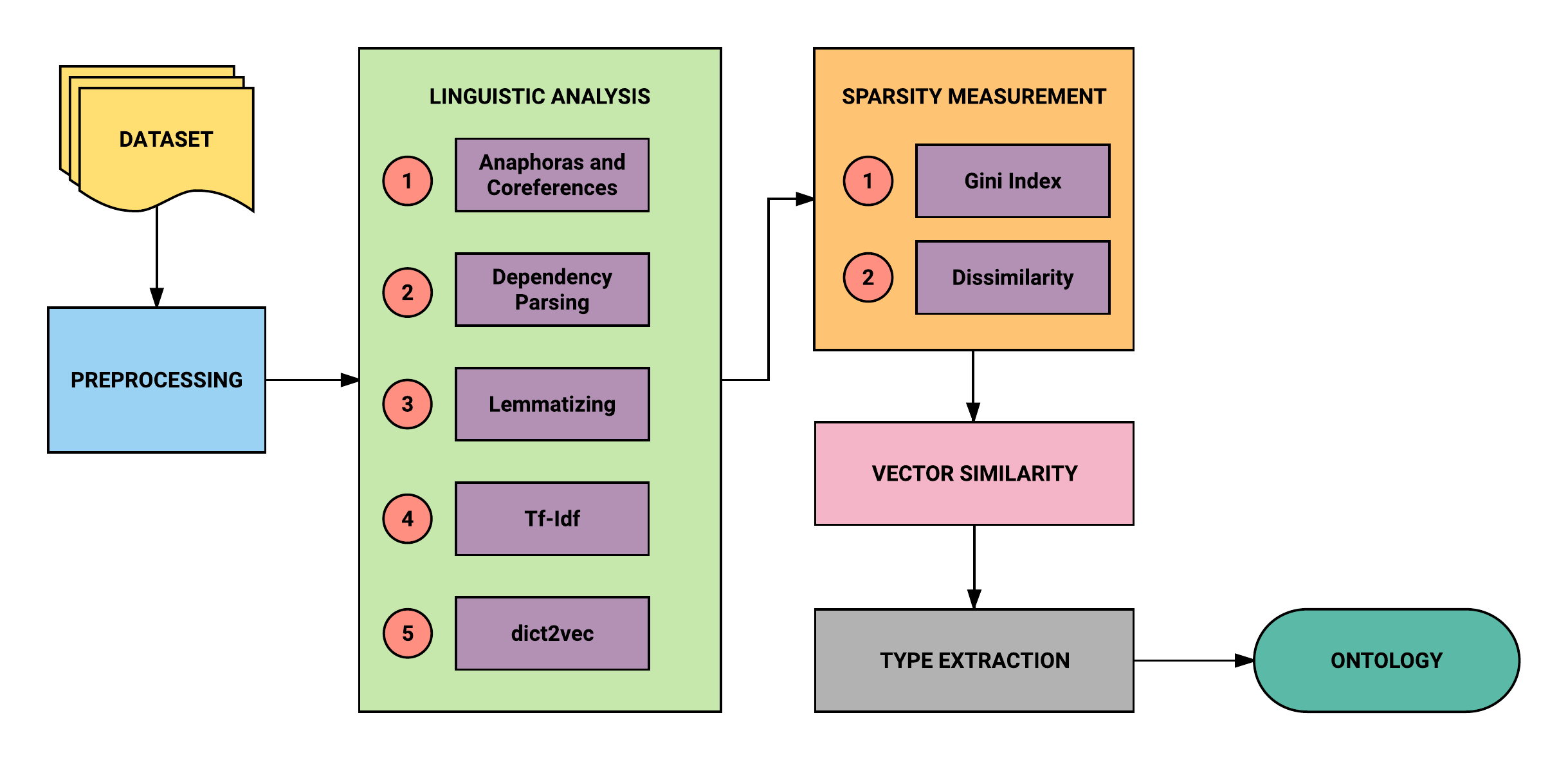}
   \caption{Workflow of the WOAH methodology.}
\end{figure}

\subsection{Data Preprocessing}

In first place, it is drastically important to correctly normalize the dataset in order to obtain tokenized sentences, by treating special words (such as hyper-links or email addresses, usually replacing them by a keyword) and removing the elements that are meaningless or provide noise to the data (special characters, \lq emoticons\rq\space or punctuation symbols that appear more than once in the same sentence). After the data is normalized, it can be divided into separated sentences in order to treat each of them in the next phase.

\subsection{Linguistic Analysis}

The second phase includes several steps, each of them facing a different problem in computational linguistics.

\subsubsection{Coreference and Anaphora Resolution}

One of the main problems in natural language processing is the resolution of coreferences and anaphoras. Coreference happens if two noun phrases refer to the same entity (e.g. "Emmanuel Macron" and "The President of the French Republic"). On the other hand, anaphora is an expression whose interpretation depends on a preceding expression in the discourse, which is called antecedent (i.e. "Go to your room and wait \textit{there}!”) (\cite{winge}, \cite{i.q.sayed}).

In order to mitigate this problem and obtain a better quality of data, a resolution model must be applied. Current state-of-the-art models apply deep reinforcement learning techniques to solve it (\cite{manning16}, for Python implementations see \cite{neuralcoref}). As the sentences have been separated, one can be passed to the module and the previous one or two can be used as the antecedent. It is important to note that although the technique does not fully remove the problem, it can considerably reduce it.

\subsubsection{Dependency Parsing}

With a better normalized data through last step, now it can be analyzed in order to obtain better results. To analyze the data, since we are interested in the extraction of verbs and nouns and their relations, a dependency parser can be used. This tool obtains the dependencies of a sentence, so if we want to extract the verbs and nouns of our dataset it can be done easily with the following process \cite{slp}:

\hfill
\begin{mdframed}[backgroundcolor=white!20]
\begin{enumerate}
\item For each sentence $s$, extract the verb $verb(s)$ and the object $obj(s)$.
\item Let $verbs$ and $objects$ be two lists and $verbsDict$ a dictionary. Also let $add2list(u, v)$ be a function that adds the element $u$ to the list $v$ and $add2dict(u, v, d)$ be a function that adds the element $u$ to the key $v$ of the dictionary $d$. Then do:
\begin{enumerate}
\item $add2list(verbs, verb(s))$
\item $add2list(objects, obj(s))$
\item $add2dict(obj(s), verb(s), verbsDict)$
\end{enumerate}
\item Obtain the nouns $comp(s)$ of $s$ that complement $verb(s)$ or complement $obj(s)$ and are common nouns.
\item Let $complements$ be a list and $objectsDict$ a dictionary. Then do:
\begin{enumerate}
\item $add2list(complements, comp(s))$
\item $add2dict(comp(s), obj(s), objectsDict)$
\end{enumerate}
\end{enumerate}
\end{mdframed}
\hfill

After that process, there will be a list with all the verbs ($verbs$), other one with all the objects ($objects$) and other with the complements that are nouns ($complements$). There should also be one dictionary ($verbsDict$) with the verbs associated to the objects and other ($objectsDict$) with the objects associated to the complements.

\subsubsection{Lemmatizing}

Now that the verbs and objects from the data have been separated, we can normalize them by reducing inflectional forms. By doing that we will remove redundant elements that add noise to our model \cite{slp}. 

Lemmatizing should be performed on the list of objects ($objects$) and the list of complements ($complements$) in order to replace plural forms. Then the list of verbs ($verbs$) should also be lemmatized to normalize the possible forms.

Finally, both dictionaries ($verbsDict$ and $objectsDict$) and the normalized dataset from step 1 (it will be used in the next step) have to be updated to reflect this normalization.

\subsubsection{Term Frequency and Inverse Document Frequency}

After stemming and lemmatizing, it is recommended to filter the verbs, objects and complements in terms of their frequency in the dataset. That is due to the fact that those elements are going to be the vectors and the dimensions of those vectors, so in order to consider an element as a dimension it has to occur in our data as much as it makes sense to consider it as relevant.

To obtain the frequencies we compute the tf-idf for each verb, object or complement. The \textbf{term frequency} ($tf_{t,d}$) is defined (\cite{ir}) as the number of occurrences of term $t$ in document $d$. The \textbf{document frequency} ($df_{t}$) is the number of documents in the collection that contain a term $t$, so with $N$ as the total number of documents in a collection, the \textbf{inverse document frequency}
($idf_{t}$) of a term t is defined as:
$$idf_{t} = \log{\frac{N}{df_{t}}}$$
Now we can define the \textbf{term frequency and inverse document
frequency} ($tf-idf_{t,d}$) as a composite weight for each term in each document:
$$tf\textit{-}idf_{t,d} = tf_{t,d} \times idf_{t}$$

In this context, for each of the three lists the $tf$\textit{-}$idf$ will be computed separately, taking $t$ as each word in a document. In the case of the verbs, the unit considered as a document $d$ is each \textbf{dialogue}. For the objects each key (a list of objects) of the verbs dictionary and finally for the complements each key (a list of complements) of the objects dictionary. $N$ is the total number of documents.

$tf$\textit{-}$idf$ is \textit{highest} when when $t$ occurs many times within a small number of documents, \textit{lower} when $t$ occurs fewer times in a document or occurs in many documents, and \textit{lowest} when $t$ occurs in virtually all documents. This means that for each list we should filter with a \textbf{threshold} ($t_{v}$, $t_{o}$, $t_{c}$) the greatest value that we want to admit as relevant. This is the first parameter that will vary depending on the specific dataset considered.

When the values of $t_{v}$, $t_{o}$ and $t_{c}$ are decided, the elements associated to each can be filtered, removing the rest (noise and stop words) from their corresponding lists and dictionaries.

\subsubsection{dict2vec}
  
After the lists and dictionaries are updated with the $tf$\textit{-}$idf$ process and the repetitions (used to obtain the frequencies for $tf$\textit{-}$idf$) can now be removed, the associations can now be converted into vectors through the following process (which we called $dict2vec$):

First for the verbs:

\hfill
\begin{mdframed}[backgroundcolor=white!20]
\begin{enumerate}
\item Let $V$ be a $n_{v} \times m_{v}$ matrix where $n_{v}$ is the number of verbs and $m_{v}$ is the number of objects, let $l.pos(a)$ denote the position of $a$ in a list $l$.
\item For each verb $v$ in $verbsDict$:
\begin{enumerate}
\item if an object $o$ from $objects$ is in the key $verb$ of $verbsDict$:
\begin{enumerate}
\item $V[verbs.pos(verb)][objects.pos(o)] = 1$
\end{enumerate}
\item else:
\begin{enumerate}
\item $V[verbs.pos(verb)][objects.pos(o)] = 0$
\end{enumerate}
\end{enumerate}
\item Normalize each row of $V$ (each verb embedding).
\end{enumerate}
\end{mdframed}
\hfill

Then for the objects:

\hfill
\begin{mdframed}[backgroundcolor=white!20] 
\begin{enumerate}
\item Let $O$ be a $n_{o} \times m_{o}$ matrix where $n_{o}$ is the number of objects and $m_{o}$ is the number of complements.
\item For each object $o$ in $objectsDict$:
\begin{enumerate}
\item if a complement $c$ from $complements$ is in the key $objects$ of $objectsDict$:
\begin{enumerate}
\item $O[objects.pos(object)][complement.pos(c)] = 1$
\end{enumerate}
\item else:
\begin{enumerate}
\item $O[objects.pos(object)][complement.pos(c)] = 0$
\end{enumerate}
\end{enumerate}
\item Normalize each row of $O$ (each object embedding).
\end{enumerate}
\end{mdframed}
\hfill

At the end of this process we obtain in $V$ the normalized embeddings for each selected verb and in $O$ the normalized embeddings for each selected object.

\subsection{Sparsity Measurement}

When we have the embeddings (with or without a dimensionality reduction), some of them will be more generic, while others more specific. In order to obtain an ontology by grouping the embeddings, a filter has to be applied in the vector space. The process will mainly consist in choosing the appropriate embeddings from which obtain others that are similar to them. Those embeddings have to be representative of the set but nor too specific (which would point to a too poor subspace), neither too generic (which would point to a too generic subspace).

To obtain those nor-too-specific neither-too-generic embeddings we should be able to determine quantitatively the level of \lq generality\rq\space associated to each embedding. It can be observed that more general embeddings are more dense and more specific ones are more sparse, so we have to determine the sparsity level of each embedding in order to filter those that have a level not very high, neither very low.

Different metrics of sparsity exist, but, as the authors evaluate and prove through different situations and properties in \cite{hurley09}, the most complete one is the \textbf{Gini index} or Gini coefficient. This metric is usually applied to economic models to measure the inequity of a distribution, which can be thought as the sparsity. The authors define a \textbf{sparse representation} as one in which a small number of coefficients contain a large proportion of the energy. In our domain, it is a \textit{representation in which a small number of dimensions (contextualizations) contain a large proportion of the possible meanings (contexts)}.

Consider a vector $x$. If their components $x_{i}$ are placed in ascending order, some of the comparisons of the original formula can be avoided leading to a less expensive computation \cite{gini}, and obtaining the following expression for the Gini index of $x$:
$$G(x) = \frac{\sum_{i=1}^{n} (2i - n - 1)x_{i}}{n\sum_{i=1}^{n} x_{i}}$$
where $x_{i}$ is an embedding dimension value in our context, $n$ is the number of values (value of a dimension in the array) observed, and $i$ is the rank of values in ascending order (the real position of the element $x_{i}$ in the embedding $x$).

Now that the similarity metric has been chosen, the Gini index is computed for each embedding. When we have the index values, the median (denoted as $m_{g,v}$ for verbs and $m_{g,o}$ for objects) of the distribution given by the Gini index obtained values is selected, which will behave as the \lq centroid\rq\space from which embeddings are filtered. The number of embeddings $\textbf{g}$ to filter is a parameter that can be adjusted to obtain more generic or more detailed results. More specifically, it determines the level of what is considered as \lq dissimilar\rq\space.

Let $dissim(a, b) = 1 - sim(a, b) = 1 - cos(a, b)$ define the \textbf{cosine dissimilarity} between $a$ and $b$. The process to filter the $g$ first embeddings (the same for both verbs and objects) is:

\hfill
\begin{mdframed}[backgroundcolor=white!20] 
\begin{enumerate}
\item Starting from the median position ($m_{g,v}$ or $m_{g,o}$), take the embedding $e$ associated to that value.
\item For each element $a$ in $e$ and each element $b$ in the list of embeddings (which can be $verbs$ or $objects$):
\begin{enumerate}
\item Compute $dissim(a, b)$.
\item Filter the $g$ dissimilar nearest embeddings $e_{f}$ ($v_{f}$ or $v
o_{f}$) of $e$
\item Insert the filtered embeddings into a list ($filteredVerbs$ for filtered verb embeddings $v_{f}$, $filteredObjects$ for filtered object embeddings $o_{f}$)
\end{enumerate}
\end{enumerate}
\end{mdframed}
%\hfill

\subsection{Vector Similarity}

From the obtained list of filtered verbs ($filteredVerbs$) and the obtained list of filtered objects ($filteredObjects$) we now have representative vectors that are not similar, so we can compute for each of them the cosine similarity with the rest in order to obtain the vectors most related with each filtered representative embedding.

The number of embeddings $\textbf{c}$ to choose, as in the case of of $g$, is a parameter that constitutes the \textbf{learning rate} of the process and can be adjusted to obtain different ranges of detail.

The process for this step behaves as follows (the same for verbs and objects):

\hfill
\begin{mdframed}[backgroundcolor=white!20] 
\begin{enumerate}
\item For each filtered embedding $e_{f}$ ($v_{f}$ or $o_{f}$) in the list of filtered embeddings ($filteredVerbs$ or $filteredObjects$) and for each $e$ ($v$ or $o$) in the list of embeddings ($verbs$ or $objects$ respectively):
\begin{enumerate}
\item Compute $sim(e_{f}, e)$
\item Select the $c$ similar nearest embeddings ($e_{s,f}$) of each filtered embedding $e_{f}$
\item Insert, along with the associated $e_{f}$, the selected embeddings into a list ($selectedV$ for selected verb embeddings $v_{s}$, $selectedO$ for selected object embeddings $o_{s}$)
\end{enumerate}
\item Insert each list of selected embeddings for a filtered one ($selectedV$ or $selectedO$) into a list (defined as $selectedVerbs$ for filtered verb embeddings $v_{f}$ and as $selectedObjects$ for filtered object embeddings $o_{f}$)
\end{enumerate}
\end{mdframed}
\hfill

Every embedding of those \lq filtered representative embeddings\rq\space $e_{f}$ constitutes an \textbf{intent type} when $e_{f}$ (i.e. $v_{f}$) is in $filteredVerbs$ or an \textbf{entity type} when $e_{f}$ (i.e. $o_{f}$) is in $filteredObjects$. This occurs because the filtered embeddings were obtained determining the level of generality, so those embeddings should have certain inherent level of abstraction. 

Furthermore, each of the selected embeddings for a filtered embedding constitutes a term associated to that type.

\subsection{Type Extraction}

When the resulting selected embeddings for each filtered embedding are generated, in order to obtain a unequivocal representation of each type, those embeddings are added along with their filtered embedding and then the resultant embedding is normalized. Each resultant verb is finally added to a matrix $I$ (of estimated intents) and each resultant object to a matrix $E$ (of resultant estimated entities).

After that process, we end up with one embedding for each type but preserving the dimensions, so the relations remain in the vectors but now we have higher-level representations which estimate an ontology for the data with the level of generalization previously specified through the parameters ($t_{v}$, $t_{o}$ and $t_{c}$ thresholds for $tf$\textit{-}$idf$, $g$ for sparsity measurement and $c$ for vector similarity).

\section{Implementation}
It is important to mention that the accuracy of this methodology will depend on the specific tools chosen for each phase.

As an example of this, currently this work is being applied to the Maluuba Frames dataset \cite{maluuba} and noise is being experienced in the results due to the mechanism of the dependency parsing tool applied, this means that each step should be evaluated during the implementation.

Finally, when the process finishes, an \textbf{evaluation} has to be performed with the resulting ontology in order to determine the accuracy of the methodology. To execute this evaluation, an appropriate \textbf{gold standard} should have been developed for the specific ontology obtained \cite{gold}. This is a large task that is being performed along with the optimization of each phase of the process.

\section{Conclusions}
WOAH applies syntactic techniques in order to extract the verbs and nouns and the relations between them by refining that information through several linguistic and statistical metrics. This methodology provides a \lq configurable-grained\rq\space ontology estimation that has as its most direct application the extraction of the approximated types of entities and types label. It also extracts their relations in function of the data and without the use of any other external source or \textit{a priori} knowledge apart from the one of the dependency parser and the module for coreferences (necessary to use their modules).

After describing the different processes, we can conclude that Weighted Ontology Approximation Heuristic is a preliminary zero-shot theoretical methodology that could be used to generate estimated ontologies of configurable levels of generalization from different data for a conversational context. 

\section{Future Work}

Lots of things can be done from this model. In first place, when the representations and relationships have been extracted, they could be used to structure the associated ontology by describing it through \textbf{Web Ontology Language} (OWL) schemes. This could also allow a visual representation.

Secondly, it can be applied to different datasets in order to study its \textbf{consistency} (since this methodology is supposed to be a way to learn from data without applying other previous information).

Furthermore, several studies can be done related to the \textbf{parameters} of WOAH to analyze if their values for a specific level of generality follow any distribution independently of the nature of the data or if it works better under certain constraints of the range of possible values.

Another objective would be to determine if the components obtained with a dimensionality reduction technique (such as \textbf{LSA}) could be mapped to specific dimensions (contexts) of the original embeddings when this reduction is necessary (in case of obtaining an excessive number of dimensions).

A previously mentioned task to advance would be to generate a gold standard to validate the result and a study of the most appropriate tools for each step of the process.

Finally, it could be used as an automatic solution for \textbf{entity and intent type extraction} and, what is more useful, to map between those types. Through that, a system that can handle ambiguities should be obtained (each possible contextualization will be normalized as a probability between 0 and 1).

%\begin{acks}
%  I would like to thank to Paulo, %Joel and Nerea their patient helps %and hopes. 
%\end{acks}

\bibliographystyle{ACM-Reference-Format}
\bibliography{sample-bibliography} 

\end{document}